\documentclass{article}

\pdfoutput=1 
\usepackage{arxiv}

\usepackage[utf8]{inputenc} % allow utf-8 input
\usepackage[T1]{fontenc}    % use 8-bit T1 fonts
\usepackage{hyperref}       % hyperlinks
\usepackage{url}            % simple URL typesetting
\usepackage{booktabs}       % professional-quality tables
\usepackage{amsfonts}       % blackboard math symbols
\usepackage{nicefrac}       % compact symbols for 1/2, etc.
\usepackage{microtype}      % microtypography
\usepackage{lipsum}		% Can be removed after putting your text content
\usepackage{graphicx}
\usepackage[numbers]{natbib}
\usepackage{doi}

\usepackage{wrapfig}
\usepackage{xcolor,colortbl}

\definecolor{LightCyan}{rgb}{0.88,1,1}

\title{KAM - a Kernel Attention Module for Emotion Classification with EEG Data}

\author{     \href{https://orcid.org/0000-0002-4862-7182}{\includegraphics[scale=0.06]{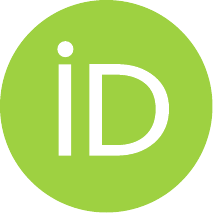}\hspace{1mm}Dongyang Kuang}\thanks{This preprint has not undergone peer review. The link to The Version of Record of this contribution can be found at \url{https://link.springer.com/chapter/10.1007/978-3-031-17976-1\_9}.} \\
	School of Mathematics (Zhuhai)\\
	Sun Yat-sen University\\
	Zhuhai, Guangdong, China \\
	\texttt{dykuang@outlook.com} \\
	\And
	\href{https://orcid.org/0000-0002-6356-233X}{\includegraphics[scale=0.06]{orcid.pdf}\hspace{1mm}Craig Michoski} \\
	Oden Institute for Computational Engineering and Sciences\\
	University of Texas at Austin\\
	Austin, USA\\
	\texttt{michoski@oden.utexas.edu}\\
}

\date{}

\renewcommand{\shorttitle}{KAM for Classification with EEG Data}

\begin{document}
	\maketitle
	
	\begin{abstract}
	In this work, a kernel attention module is presented for the task of EEG-based emotion classification with neural networks. The proposed module utilizes a self-attention mechanism by performing a kernel trick, demanding significantly fewer trainable parameters and computations than standard attention modules. The design also provides a scalar for quantitatively examining the amount of attention assigned during deep feature refinement, hence help better interpret a trained model. Using EEGNet as the backbone model, extensive experiments are conducted on the SEED dataset to assess the module's performance on within-subject classification tasks compared to other SOTA attention modules. Requiring only one extra parameter, the inserted module is shown to boost the base model's mean prediction accuracy up to more than 1\% across 15 subjects. A key component of the method is the interpretability of solutions, which is addressed using several different techniques, and is included throughout as part of the dependency analysis. 
	\end{abstract}

	% keywords can be removed
	
	\section{Introduction}
	Correctly identifying human emotion using classification strategies has long been a topic of interest in brain computer interfaces (BCI) and their applications. According to the review \cite{lotte2018review} on classification algorithms utilized in EEG studies, there are five major categories of classifiers currently under investigation, which are: i) conventional classifiers \cite{schlogl2009adaptive,li2010bilateral,liu2010improved,liu2012unsupervised}, ii) matrix and tensor based classifiers \cite{congedo2017riemannian}, iii) transfer learning based methods \cite{blankertz2008invariant,fazli2009subject}, iv) deep learning algorithms and advanced statistical approaches \cite{cecotti2010convolutional,lu2016deep}, and v) multi-label classifiers  \cite{lotte2007review,blankertz2011single,steyrl2016random}. While many classification approaches have been explored in the context of EEG signal processing, the classification pipeline itself has still frequently involved extensive manual preprocessing and feature engineering, often requiring intervention from domain experts as well as the experimental operators involved in the acquisition of the EEG signals themselves. %However ... 
	
	The success of neural network tools has been shown to alleviate some of limitations in the classical pipeline by providing, for example, faster predictions and reducing the need to manually preprocess data.  One problem that remains however, is that the scale of learnable parameters in a classification network can be too large relative to the input data size. That is, in contrast to areas such as image classification, the availability of clean and open-sourced EEG data sets is comparably quite small in size. As a consequence, basic research calls specifically for data efficient and parameter efficient, as well as human interpretable models in order to provide penetrating insight into human emotion classification given relatively sparse data sampling.
	
	\section{Related Work}
	Below we briefly review some salient results in the literature relevant to the present work.
	
	\begin{wrapfigure}{l}{0.3\textwidth}
%		\vskip-0.75cm
		\includegraphics[width=0.3\textwidth]{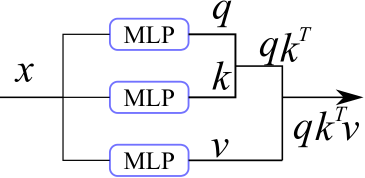}
		\caption{The basic self-attention mechanism.}\label{fig:self-attention}
%		\vskip-0.75cm
	\end{wrapfigure}
	\textbf{Self-Attention:} Self-attention was originally introduced in the field of natural language processing (NLP) \cite{bahdanau2016neural}, where the self-attention mechanism operates as a key component in transformer modules. These standard self-attention designs rely on MLPs, instead of the more conventionally used convolutional layers, to generate an attention matrix. Adapting the concept of self-attention to computer vision has been remarkably successful, and has demonstrated impressive performance to date on a variety of different tasks, e.g. \cite{dosovitskiy2020image,liu2021swin}. As a consequence, through self-attention approaches, the field of image classification--- long dominated by an assortment of solutions utilizing convolution neural networks---has discovered a promising new tool for (potentially) broad application.
	
	\textbf{Alternative Attention Approaches:} In addition to self-attention, there exists many other types of attention modules in the field of computer vision. Two prominent examples of these are the Squeeze-and-Excitation (SE) network \cite{hu2018squeeze} and the Convolutional Block Attention Module (CBAM) \cite{woo2018cbam}, where the SE network won the ImageNet2017 championship, while CBAM sequentially infers attention maps along both channel and spatial dimensions for adaptive feature refinement.  These methods, and others, are indicative of the increased contemporary importance  attention-style approaches are having within computer vision.
	
	\textbf{EEGNet:} EEGNet was proposed in \cite{lawhern2018eegnet} for a compact network design aimed at finding better generalizations across different BCI paradigms. Using depthwise/separable convolution layers, the EEGNet network contains considerably fewer trainable parameters than models constructed with regular convolutional layers, while still showing commensurate performance. EEGNet is also equipped with a convolutional layer of kernel size equal to the total number of EEG channels, enabling the ability to investigate spatial patterns learned during training in order to elucidate underlying electrophysical principles. Because of both the effectiveness and parsimonious nature of the EEGNet design, we adopt it as the backbone network for comparing and interpreting different attention modules.
	
	\textbf{SEED:} The SEED dataset \cite{zheng2015investigating} includes multiple physiological signals that evaluate self-reported emotional responses, classified into {\it Positive}, {\it Neutral}, and {\it Negative} reactions taken from 15 participants, and is one of the standard datasets for benchmarking EEG signal classification strategies. The data was collected with 62 EEG channels using the 10–20 international standard. %The collected EEG signals were manually preprocessed and further downsampled to 200 Hz followed by a bandpass filter of 0-75 Hz. %The performance of many different network designs have been tested on the SEED dataset, with extensive work being published in the literature. %For example, the first performance reported on SEED was from \cite{zheng2015investigating} with a best overall accuracy of 86.65\%. In \cite{zhang2018spatial} a STRNN design raised was able to improve that number to 89.5\%. With the help of the DGCNN architecture, authors of \cite{song2018eeg} reported a 90.4\% mean accuracy. This accuracy was then further raised in \cite{li2018novel}, where a BiDANN model achieved 92.38\% accuracy.  In \cite{asghar2019eeg}, a ``bag of deep features'' was used to achieve an accuracy of 93\% with a well-tuned support vector machine (SVM) classifier, and this was further improved in \cite{li2019regional} to a mean accuracy of 93.38\% using a R2G-STNN model.   
	
	\textbf{Our work:} In this paper we consider a self-attention mechanism for boosting the backbone model's performance in a parameter efficient way and for providing better interpretation. However, at the outset, it is worth noting that several potential difficulties arise when conceptualizing the incorporation of a self-attention mechanism into the EEGNet framework.  First, as discussed in \cite{dosovitskiy2020image}, transformers tend to be quite data hungry models, failing to outperform regular convolution-based networks when the dataset's size is not large enough. In the area of emotion classification using EEG signals, this data-thirst requirement can become prohibitive. One of the reasons for this ``data hungry'' aspect of transformers is due to the MLP layers generally involved in the attention mechanism. These dense layers, not surprisingly, tend to contain many more parameters than convolutional layers that comprise popular emotion classification frameworks.  Consequently, a difficulty arises when self-attention is applied directly to frameworks such as EEGNet, since the resulting hybrid frameworks tend to substantially undermine the primary advantage of the underlying base models; for example, EEGNet would no longer be a lightweight and compact model, but instead become a data hungry model focusing entirely on accuracy over pragmatic utility. 
	
	Thus, the primary goal of the present work is to find a way to incorporate the self-attention mechanism into EEGNet in a way that can still preserve EEGNet's pragmatic utility by maximizing the parameter efficiency in the design of the attention module.  The solution presented in this paper is called a Kernel Attention network Module (KAM), and can be described as:
	\begin{enumerate}
		\item Utilizing a kernel function to produce the proper attention matrix instead of relying on MLP layers; thus reducing the number of both parameters and computations required.  Moreover, benefiting from the one parameter design, specific techniques are then able to be employed for more effective interpretability techniques as well.
		\item With the proposed module inserted along with only one additional parameter, the predictive performance of the baseline model---in our case, EEGNet---can be boosted up to more than 3\% for some subjects on within-subject classifications and more than 1\% overall on mean performance. 
	\end{enumerate}
	
	\section{Kernel Attention Module}\label{sec:KAM}
	
	Figure \ref{fig:self-attention} gives a simple illustration of how the basic self-attention mechanism works when applied on some feature $x$. First, the feature $x$ is mapped to three different features of the same size via $q = \phi_q(x)$, $k = \phi_k(x)$, and $v = \phi_v(x)$. The mappings $\phi_j$ for $j\in\{q,k,v\}$ are achieved using MLP blocks. Next, an attention matrix is formed by computing $qk^T$ which will then be used as a prefactor on $v$ to produce the attention output. Computationally, this procedure can be summarized as: $x \leftarrow [\phi_q(x)\phi_k(x)^T]\phi_v(x)$.
	
	\begin{wrapfigure}{r}{0.35\textwidth}
%		\vskip-0.45cm
		\includegraphics[width=0.35\textwidth, height=0.45\textheight, trim = 0 0 0 0, clip]{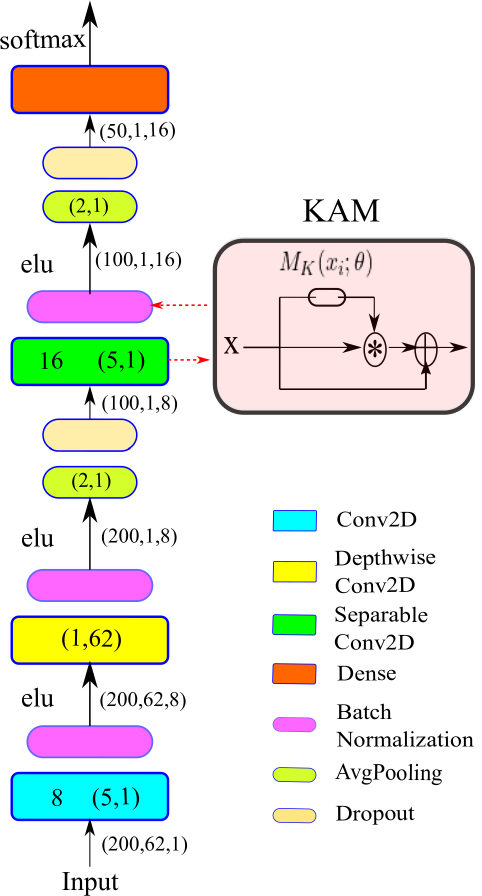}
		\caption{EEGNet with KAM inserted. Some important hyperparameters, kernel shapes and tensor sizes are also shown.}\label{fig:KAM}
		% 	\vskip-0.4cm
	\end{wrapfigure}
	In the case where $x$ has large feature dimension $n$, the attention is applied on segments of $x$ in parallel with the resulting feature pieces subsequently concatenated afterwards---a mechanism referred to as \emph{multi-head self-attention}. For more details on the underlying algorithms we refer the reader to \cite{dosovitskiy2020image}. Using the above basic self-attention format, KAM is constructed by replacing the inner product form $\phi_q(x)\phi_k(x)^T$ with a kernel matrix $M_K(x;\theta)$ subject to some parameter $\theta$.  For example,  a Gaussian type kernel function can be used to generate $M_K^{ij} = \exp( -\alpha d( x_i, x_j)^2 )$, where $d(\cdot, \cdot)$ denotes some distance metric, $x_i$ is the $i$th row or column of feature block $x$ depending on whether $M_K$ is multiplied to $x$ by left or right, $\theta=\alpha$ is the learnable parameter during training. In the KAM design, $\phi_v$ can be simply dropped to reduce the number of total parameters. Finally, a skip connection is included in the the KAM design that offers several potential benefits. On one hand, an additional skip can help better backpropagation of gradients to the blocks in front of the KAM. On the other, it provides an easy interface to quantitatively measure how much attention is actually being applied, requiring only an examination on the values of $\theta$. For example, when $\alpha \rightarrow +\infty$ then $M_K(x;\theta) \rightarrow I$, meaning no cross attention among features is applied. However, if $\alpha \rightarrow 0$, then $M_K(x;\theta) \rightarrow J$ which is an attention matrix whose off-diagonal entries $J_{ij} = 1, i\neq j$, meaning deep features now equally contribute to others for refinement during training. The above procedure leads to our Kernel Attention Module (KAM) design as shown in Figure \ref{fig:KAM}, where its symbolic form can be summarized as:
%	\vskip-0.3cm
	\begin{equation}\label{eqn:KAM}
		x \leftarrow x + M_K(x;\theta)x = \left(I + M_K(x;\theta)\right)x.
	\end{equation} 
	% \vskip-0.3cm
	
	The proposed KAM mechanism can also be easily applied with multiple heads. We further note that in the implementation in Section \ref{sec:exp}, an extended form is used for $M_{K}^{ij} = \exp(-\alpha d(x_i, x_j)^2)$, where $\alpha \in (a, \infty)$. If the lower bound is set to $a=0$, the case can be readily interpreted as $\alpha = 1/{{\sigma}^2}$ where the parameter can be understood as a kind of "standard deviation".  Setting $a$ less than zero allows  off-diagonal entries in $M_K$ to have values greater than one, in which case $a$ should be close to zero, i.e. $a\approx -\epsilon$ for $\epsilon$ small, to prevent numerical blow-up during training from poor matrix conditioning. 
	
	\section{Experiments}\label{sec:exp}

\begin{wrapfigure}{l}{0.45\textwidth}
	%		\vskip-0.5cm
	\includegraphics[width=0.45\textwidth, height=140pt, trim = 0 0 0 0, clip]{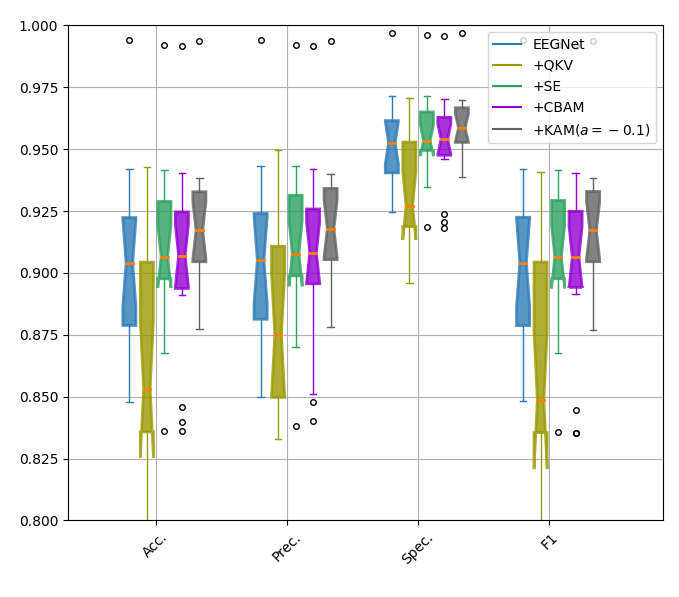}
	\caption{Overall mean prediction performance across 15 subjects.}\label{fig:box}
	%		\vskip-0.5cm
\end{wrapfigure}

	In this research, we focused on model's performance on subject dependent classification tasks. In the spirit of \cite{zheng2015investigating} the data set was divided into non-overlapping epochs, each lasting one second, yielding $\sim$3300 epochs per trial, per subject, and $\sim$1060 epochs per labelled emotion. However, in contrast to \cite{zheng2015investigating}, and most other studies where the training and testing data are manually split and models evaluated in a single pass, we adopt a cross-validation (CV) approach to improve evaluation robustness. 

	In the following benchmark using EEGNet with KAM, the data from each subject is split, taking 1/6 for validation during training. Five-fold cross validation (5-CV) is then performed on the remaining 5/6 of the data. Theses ratios are chosen to make the validation and test set roughly the same size during cross validation. For any model test, initial weights are set to the same values across each of the five folds. The network is trained for each fold over a maximum of 80 epochs, and the best model is selected at the epoch with the best validation accuracy. All experiments are trained with the same Adam optimizer configuration of an initial learning rate of $10^{-2}$ and a decay rate of $0.75$, which only activates when no accuracy improvement is seen on the validation set in the past 10 epochs. A total of $5\times15 = 45$ training runs are conducted for each model compared in our benchmark. The code used for models' training and evaluation will be made available at \url{https://github.com/dykuang/BCI-Attention}

\begin{figure}[tbh]
		% 	\vskip-0.3cm
		%   \includegraphics[width=0.45\textwidth]{sigma_dist.png}
		%   \includegraphics[width=0.45\textwidth]{sigma_vs_acc.png}
		%   \includegraphics[width=\textwidth]{dependence_alpha.eps}
		\includegraphics[width=\textwidth]{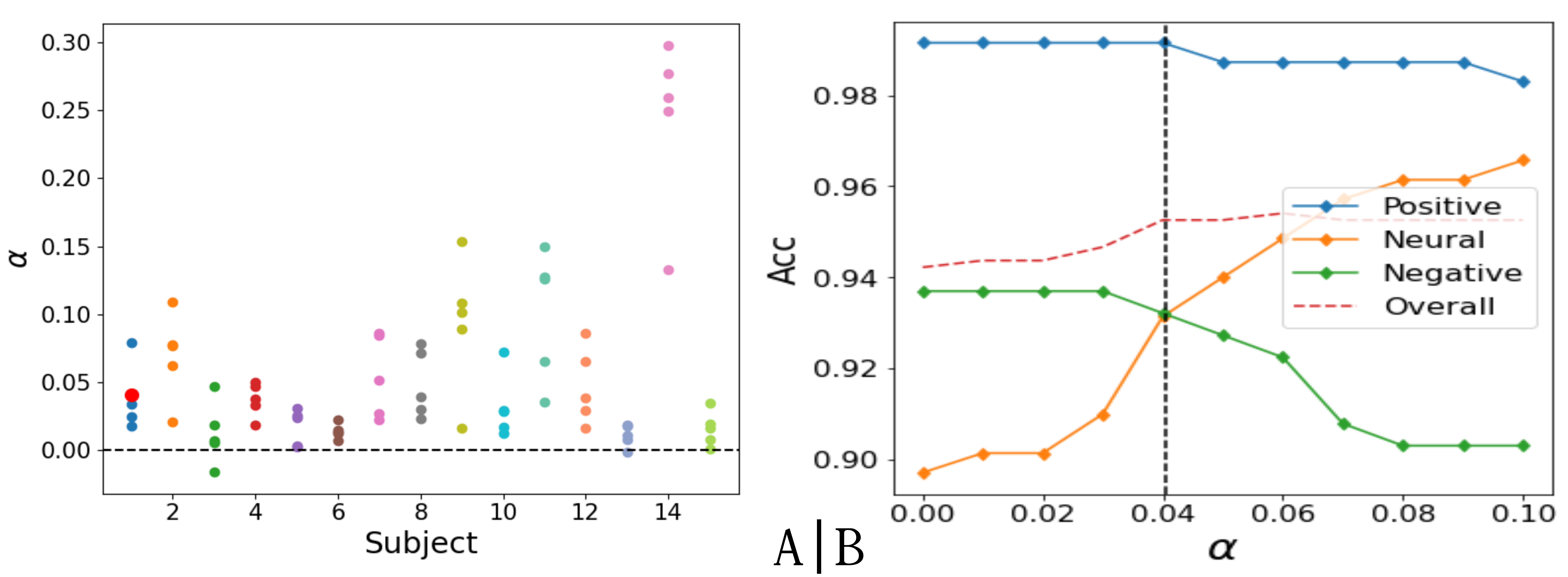}
		\caption{%$A|B$ \CEMnote{This is confusing, can we add A and B to the plots?}, 
			$A$: Distribution of learned $\alpha$ value during the 5CV with EEGNet+KAM across the 15 subjects. $B$: Change of accuracy with varying value of $\alpha$ while freezing other parameters in the selected model (marked as red in first column i.e. subject S01 of $A$).}\label{fig:sigma}
		% 	\vskip-0.5cm
	\end{figure}

	\textbf{Benchmark:} For benchmarking we compared five models: a) EEGNet, b) the basic $QKV$ type attention from Figure \ref{fig:self-attention}, c) SE attention, d) CBAM attention, and e) KAM($a = -0.1$). All implementations herein are inserted at the same location shown in Figure \ref{fig:KAM}. Note that the basic QKV attention module does not perform well here, which is likely due to it, as mentioned in \cite{dosovitskiy2020image}, being data thirsty and SEED not being a large enough dataset to quench.  It is also worth mentioning here that the version of KAM with $\alpha$'s lower bound $a = 0$ gives mean accuracy of 91.74\% $\pm$ 3.02\% which is slightly worse than the case of $a = -0.1$. This suggests that the extension of the lower bound to a negative value can potentially help during deep feature refinement. We also observe in our experiments that the training procedure for some subjects will push $\alpha$ slightly below zero for minimizing the loss function (see Figure \ref{fig:sigma}).
	
	\begin{table}[tbh]
%		\vskip-0.3cm
		\caption{Mean accuracy reported from different models. %First row contains some results from previous research. Note that their numbers reported are not from 5CV experiment setting and are with much larger networks in terms of number of parameters. 
		}\label{tab:acc}
		\centering
		\begin{tabular}{c|c|c|c|c|c}
			% 		\hline
			% 		Models & STRNN\cite{zhang2018spatial} & BoDF\cite{asghar2019eeg} &  BiDANN \cite{li2018novel} &DGCNN\cite{song2018eeg} &R2G-STNN \cite{li2019regional}\\
			% 		Acc(\%) & 89.50 \tiny{$\pm$ 7.63}& 93.80 $\pm -$ & 92.38 \tiny{$\pm$ 7.04}& 90.40\tiny{$\pm$8.49} &93.38\tiny{$\pm$5.96}\\
			% 		\hline
			\hline
			\rowcolor{LightCyan}Models & EEGNet & +QKV & +SE & +CBAM & \cellcolor[HTML]{ffea00}+KAM($a=-0.1$) \\
			\rowcolor{LightCyan} Parameters &  3851 & 4940 &  3933 & 4033 & \cellcolor[HTML]{ffea00}3852\\ 
			\rowcolor{LightCyan}Acc(\%) & 90.34 \tiny{$\pm$ 3.69} & 86.81 \tiny{$\pm$4.19}& 91.20 \tiny{$\pm$3.42} & 90.40 \tiny{$\pm$3.97}&\cellcolor[HTML]{ffea00}91.89  \tiny{$\pm$2.76}  \\
			\hline
		\end{tabular}
%		\vskip-0.5cm
	\end{table}
	
	% Figure \ref{fig:improve} summarizes the mean accuracy improvements per subject with KAM inserted. Assuming independence, the plot shows the value $m_{-} = m_1-m_2$ and the standard deviation shown in red estimated by $\sigma_{-} = \sqrt{\sigma_{1}^{2}+\sigma_{2}^{2}}$, where $m_1, m_2$ are mean values and $\sigma_{1}, \sigma_{2}$ are standard deviations estimated from the 5-CV results for the two models. KAM insertion shows visible improvement on 13 of the 15 subjects tested (except for S04 and S15). Among these, improvements on S02, S05, S06-S08, S13 and S14 are more than 1$\sigma_{-}$.  
	% \begin{wrapfigure}{r}{0.6\textwidth}
	% 	\vskip-0.3cm
	% 	\includegraphics[scale=0.5, trim = 10 10 10 20, clip]{improve_acc.eps}
	% 	\caption{Estimation of $m_{-}$ and $\sigma_{-}$ for the difference per subject w or wrt KAM inserted.}\label{fig:improve}
	% 	\vskip-0.75cm
	% \end{wrapfigure}
	\textbf{Channel Attention:}  The inserted KAM module can potentially change the kernel weight originally designed in EEGNet during the model's decision process. Particularly, kernel weights in the first depthwise convolution layer (see Figure \ref{fig:KAM}) were treated as a representation for relative attention across different channels\footnote{These are kernel weights in the first depthwise convolutional layer. The shape is of(1, 62) and can be directly associated with the 62 EEG sensor locations on scalp.
	} in \cite{lawhern2018eegnet}. They can be affected when different attention modules are inserted. As the depthwise convolution applies one kernel to each of the eight channels, there are eight kernels associated to the architecture in Fig.~\ref{fig:KAM}. For clarity here, to illustrate the effect of these kernels on the network, we only examine the kernel applied on the first feature channel. 
	
	\begin{figure}[tbh]
%		\vskip-0.5cm
        \centering
		\includegraphics[width=\textwidth, trim=0 0 0 0, clip]{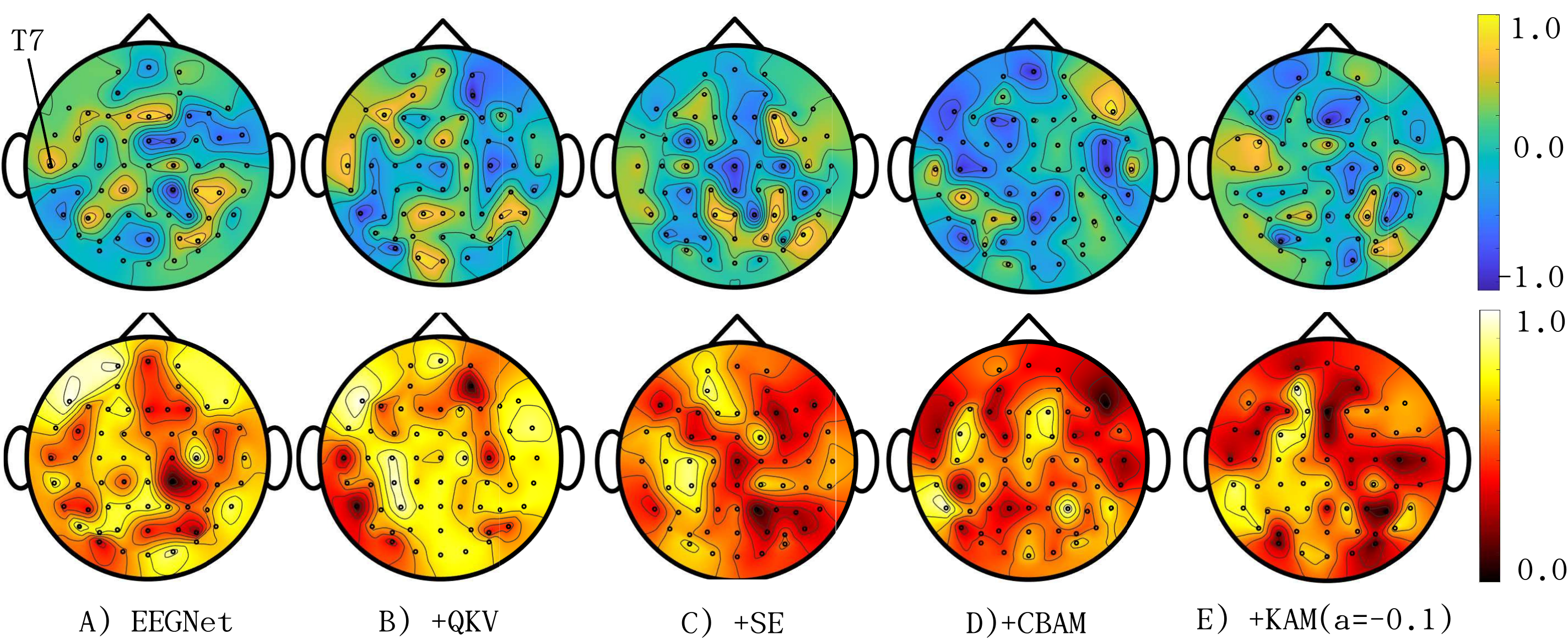}
		\caption{Kernel weights mapped onto scalp maps. The first row shows the normalized mean.  The second row shows the normalized standard deviation from the 5-CV.}\label{fig:scalp}
%		\vskip-0.2cm
	\end{figure}
	
	To visualize the different spatial attention patterns discovered represented by the selected kernel weights of different models, we present scalp maps in Fig.~\ref{fig:scalp}. These maps are generated by training the models under 5-CV for subject S01, where mean and standard deviations are shown. It can be clearly seen overall that different modules can result in different channel attention patterns.  While it may be difficult to immediately associate the mean value mappings to informative clinical interpretations, the spatial magnitude of the standard deviation does provide a way to measure different models' confidence in assigning kernel weights across different regions. For example, one thing to observe here is that all models visualized high mean attention values around the T7 region with relatively low uncertainty (represented by the std value). This observation seems to support some studies reporting correlations between emotional deficiency and memory development with specific temporal lobe function, such as in the diathesis of schizophrenia, e.g. \cite{goghari2011temporal}, and in types of memory enhancement in forms of dementia, e.g. \cite{kumfor2014frontal}. As an alternative research direction, how to enforce one's prior knowledge on task related scalp patterns so that the posterior learned patterns are robust to inserted attention modules is also important for building a better human interpretable model.
	%how confident the model think kernel weights in a particular region correspond to an optimal value.
	%differences between models trained from different folds. In other words, how different a model will change its mind in making decisions according to data contained in different channels. The lower the value (red region) on certain channels, the more certain model``thinks"  the corresponding weights should be in order to get the optimum prediction from training. 

	\textbf{Dependency on $\alpha$:} For better interpret the effect of module parameter $\alpha$ from KAM in trained models, we organised this section. Figure \ref{fig:sigma}(A) shows the distribution over learned $\alpha$ during 5-CV for each subject. Among these, only experiments with data from subject S02 and S13 yield instances where $\alpha <0$, while all other trainings find $\alpha \ge 0$. However, per subject speaking, the change in the lower bound on $\alpha$ does have a noticeable impact. For example, in the data from subject S05, S06, and S13 the learned $\alpha$ values cluster at locations close to zero, but we observed from our experiments that this small deviation from zero results in noticeable accuracy differences. This may because of the fact that small $\alpha$ will correspond to the case with large $\sigma$, i.e. further away from zero attention as explained above in Section \ref{sec:KAM}.
	
	The one-parameter KAM design also makes it easy to analyze prediction accuracy as a function of the module parameter $\alpha$. As an example, we choose the model trained from the first fold (marked with a red dot in Fig.~ \ref{fig:sigma}(A)) for subject S01 and gather data from three trials each with a different emotion label for the dependency analysis. By varying $\alpha$ values in KAM while keeping other model parameters frozen, we can examine how $\alpha$ conditionally effects the prediction. In \ref{fig:sigma}(B), it is interesting to see that the learned value (black vertical line) in KAM happens at a location where the overall accuracy line first rises to stabilize in this case. It is also interesting to observe that crossover between accuracy lines for ``neutral" and ``negative" happens at the same location of learned $\alpha$. \footnote{This might be an interesting coincidence since we also had other cases in our experiments where they do not meet exactly.}.
	%and . 
	
\begin{figure}[tbh]
		%   \includegraphics[width=0.45\textwidth]{baseline_parital_hist.png}
		%   \includegraphics[width=0.45\textwidth]{parital_sigma_p.png}
		%   \includegraphics[width=0.45\textwidth]{partial_sigma_m.png}
		%   \includegraphics[width=0.45\textwidth]{partial_sigma_n.png}
		% 	\vskip-0.70cm
		%   \includegraphics[height = 200pt, width=\textwidth]{dependence_p_alpha.eps}
		\includegraphics[height = 250pt, width=\textwidth]{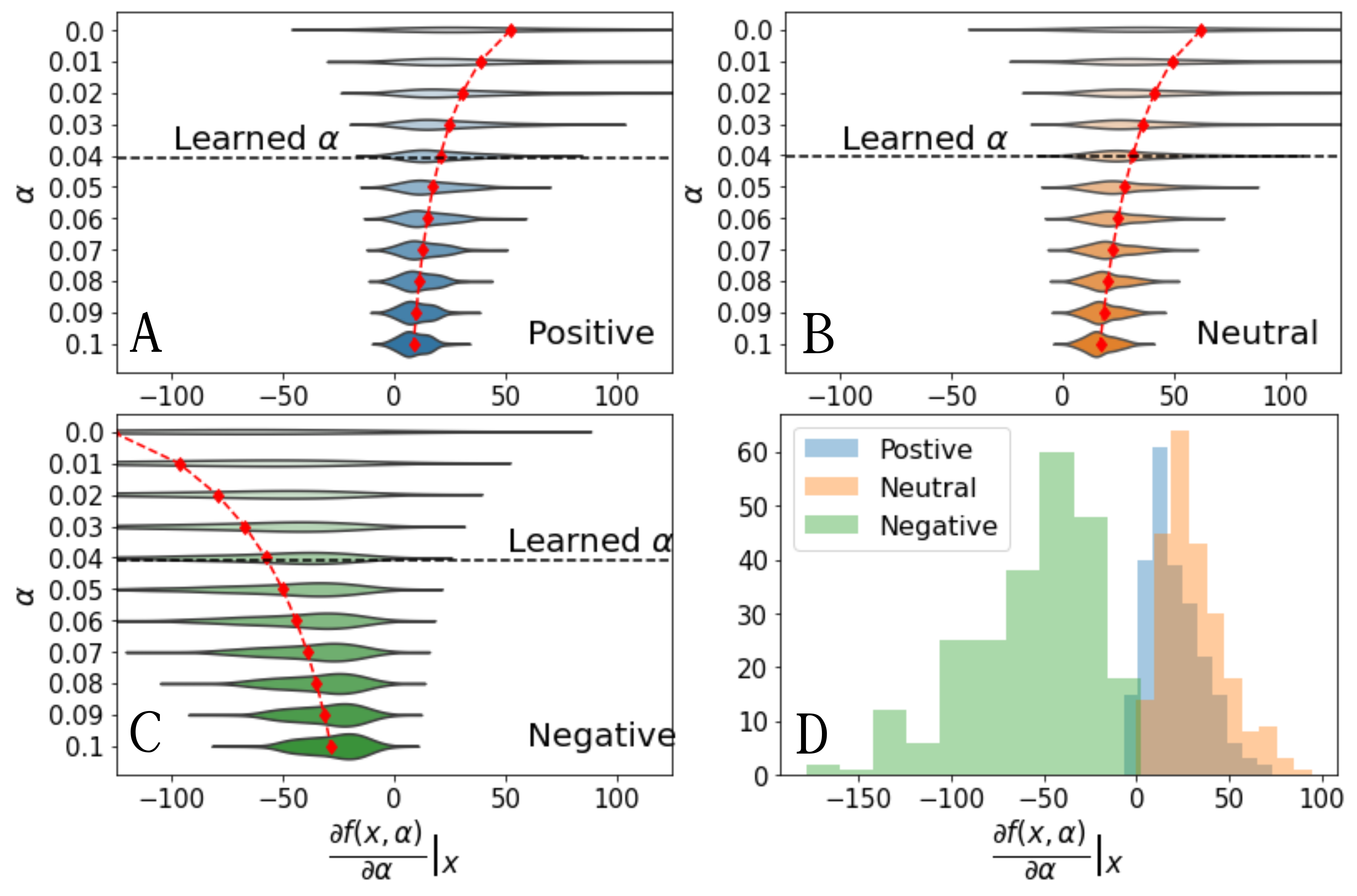}
		\caption{\textbf{A-C}: Distribution at different $\alpha$ values corresponding to label ``Positive", ``Neutral" and ``Negative". The mean is linked by dashed line. \textbf{D}: Distribution at learned $\alpha = 0.0406$ for different emotion labels. 
			%\CEMnote{I love this result, but these plots are very hard to read, the labels and text are distorted and patchy. I cannot even read the $x$-axis.}
		}\label{fig:partial}
		% 	\vskip-0.6cm
	\end{figure}

	This dependency can also be examined via the distribution of $\frac{\partial f_i(x)}{\partial \alpha}|_{x}$ for varying $\alpha$, where $x$ is the input and $f_i(x)$ is the output of the corresponding neuron (before activation) from the last dense layer for label $i\in\{1,2,3\}$, i.e. positive, neutral, and negative emotion labels respectively. The result is gathered in Fig.~\ref{fig:partial}(A-C) computed with the same data mentioned in the previous paragraph. Of note, the partial dependencies appear to show very similar patterns. That is, the variance in each is a decreasing function for $\alpha \in [0, 0.1]$. This can be explained by the fact that $\alpha$ is packaged inside an exponential form that maintains its character through differentiation. Finally, Fig.~\ref{fig:partial}(D) shows a close look at the histogram of how these distributions differ at the learned $\alpha=0.0406$ from the model's selection during training. %Partials for neutral are mostly positive and partials for negative are mostly negative, which matches with the observations on Figure \ref{fig:sigma}(B) that accuracy on neural is increasing and accuracy on negative is decreasing at the learned $\alpha$. 
	
	% a particular segment when predicted label is changed with sigma value.
	
	\textbf{Prediction Transition Curve:}
	In this section, we would like to explore how different models react when the input sample to be predicted is gradually transformed to another sample via some morphing operation $g$. In other words, we try to examine and interpret how different models react under a particularly selected ``attack" $g$. Let $\{x_0, x_1, x_2\}$ denote 3 samples with labels 0, 1 and 2, and $\vec{p_i} = \{p_i^j\} = F(x_i)$ with $j = 0, 1, 2$ denoting model $F$'s confidence (here using a softmax score of the last dense layer) assigning input $x_i$ for label $j$. Then clearly by construction $\sum_{j} p_i^j = 1$ for any $i$. Further let $g_i^j(u), u \in [0, 1]$ denote an morphing operation between samples parameterized by $u$ such that $x_i = g_i^j(0)*x_i$ and $x_j = g_i^j(1)*x_i$. The symbol $*$ here stands for certain abstract action operation. In trinary classification tasks, as $u$ increases, $F[g_i^j(u)]$ will draw a curve in the hyperplane $x+y+z=1$ inside the triangle formed by $[1,0,0], [0,1,0], [0,0,1]$. By checking these curves (we call them \textit{prediction transition curves, PTC}), one can have an idea that how the trained model $F$ reacts with respect to the morphing operation $g$ on selected samples. The idea can be generated to higher dimensional cases with class categories $n>3$, but it then becomes harder to visualize these cases as simple curves being embedded in higher dimensional simplices. This prediction transition curve provides a way for visualizing and interpreting model's predicting behavior under ``attack" $g$ for given inputs.
	\begin{figure}[tbh]
		\centering
		\includegraphics[height = 120pt, width=0.75\textwidth, trim = 10 0 10 0, clip]{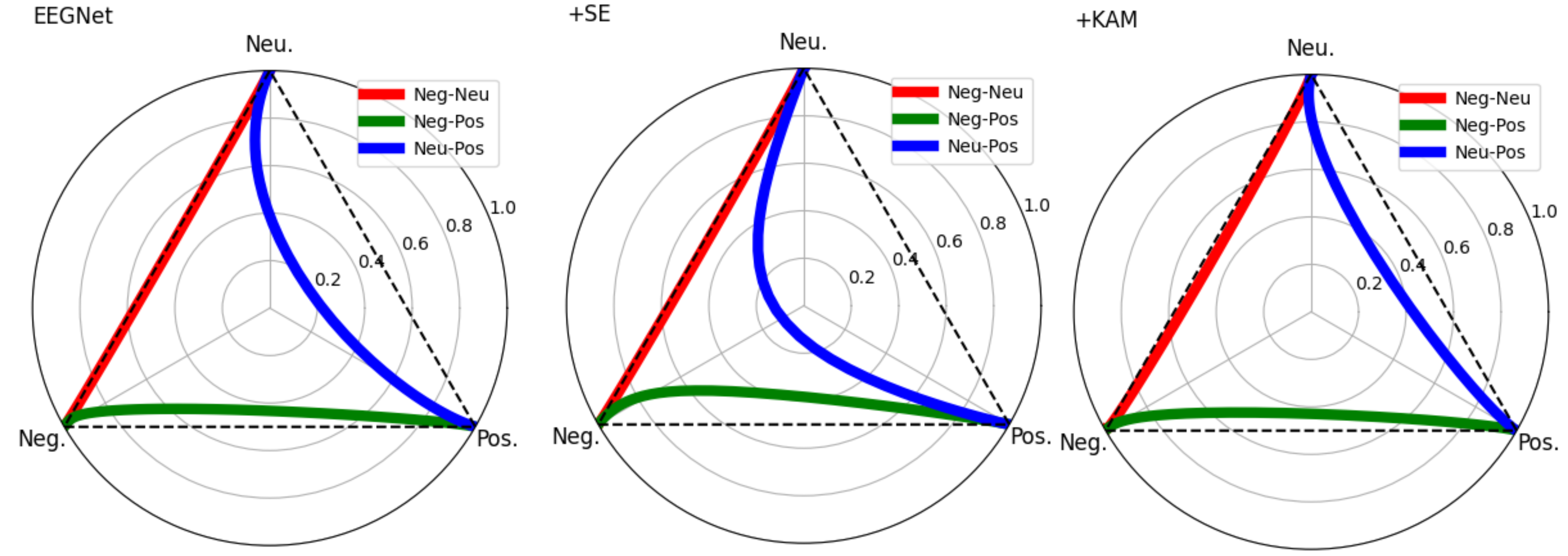}
		\caption{The prediction transition curves from the three models on the same selected samples. From left to right: EEGNet, EEGNet+SE, EEGNet+KAM.
		}\label{fig:corrupt}
		% 	\vskip-0.5cm
	\end{figure}
	
	As a demonstration, we select three samples each with a different label, and EEGNet, EEGNet+SE, EEGNet+KAM all predict correctly on them. For simplicity, we choose the straightforward linear interpolation between samples for $g$, i.e. $g_i^j(u)*x_i = (1-u)x_i + ux_j$. Notice that this definition of $g$ is symmetric in terms of $g_i^j(u)*x_i = g_j^i(1-u)*x_j$. So morphing from $x_i$ to $x_j$ and $x_j$ from $x_i$ end up with the sample path (regardless of the direction). Other types of morphing operations can also be used depending on one's prior on the (known or inferred) underlying relationships between samples. The prediction transition curves obtained from morphing with $g$ are summarized in Figure \ref{fig:corrupt}. It can be seen that all three compared models have almost straight PTCs for the connection of "Negative- Neutral" and "Negative-Positive", meaning that when one input sample is slowly morphing to the other, the model transits its confidence between the two labels almost linearly while leaving the third label almost untouched. Curves linking "Neutral-Positive" are all curved to the center at different extents, suggesting that models are in some sense "hesitating" to assign "Negative" for intermediate samples generated by morphing between "Neutral" and "Positive". This example is consistent with the observations in Figure \ref{fig:partial} D) and provides a different angle suggesting that the trained models find it more difficult to separate "Positive" from "Neutral" emotions than separating "Negative" from "Positive" or "Neutral" from "Negative" emotions for the subject data under examination.  An interesting followup question is whether this observation bears clinical significance as well, something which undoubtedly deserves consideration. 
	
	\section{Conclusion}
	In this work, we present a kernel attention module that can be inserted into a network for deep feature refinement. Using EEGNet as the backbone model, the performance of KAM are benchmarked against several SOTA attention modules under cross validation with SEED dataset. With only one additional parameter, the idea behind KAM has demonstrated good potential for developing parameter efficient models that can simultaneously help human interpretation on trained models. Many follow-up studies are possible in this context, including investigating the effects of different kernels (other than Gaussian) alongside more exhaustive dependency analyses. Additionally, examining different training strategies, such as the masked-autoencoder discussed in \cite{he2021masked}, might also be beneficial.
 
	\section{Acknowledgement}
	\textit{This work was partially supported by the Fundamental Research Funds for the Central Universities, Sun Yat-sen University (22qntd2901)}
	
	\bibliographystyle{abbrvnat}
	\bibliography{IMIMIC2022_arxiv_ref}

\end{document}